\documentclass[runningheads]{llncs}
\usepackage{graphicx}
\usepackage{amsmath,amssymb, gensymb}
\usepackage{bm}
\usepackage{subcaption}
\usepackage{subfloat}
\usepackage{float}
\usepackage[inline]{enumitem}
\usepackage{xcolor}
\usepackage[hyphens]{url}

\begin{document}
\title{Exploiting Transitivity Constraints for Entity Matching in Knowledge Graphs\thanks{Supported by the Netherlands Organisation for Scientific Research}}
\titlerunning{Entity Matching in Knowledge Graphs}
%
%
\author{J. Baas\inst{1}\orcidID{0000-0001-8689-8824} \and
M. M. Dastani\inst{1}\orcidID{0000-0002-4641-4087} \and\\
A. J. Feelders\inst{1}\orcidID{0000-0003-4525-1949}}
\authorrunning{J. Baas et al.}
%

\institute{Utrecht University, Heidelberglaan 8, 3584 CS Utrecht, Netherlands \and
\email{\{j.baas,m.m.dastani,a.j.feelders\}@uu.nl}}

\maketitle              
\begin{abstract}
The goal of entity matching in knowledge graphs is to identify entities that refer to the same real-world objects using some similarity metric. The result of entity matching can be seen as a set of entity pairs interpreted as the same-as relation. However, the identified set of pairs may fail to satisfy some structural properties, in particular transitivity, that are expected from the same-as relation. In this work, we show that an ad-hoc enforcement of transitivity, i.e. taking the transitive closure, on the identified set of entity pairs may decrease precision dramatically. We therefore propose a methodology that starts with a given similarity measure, generates a set of entity pairs that are identified as referring to the same real-world objects, and applies the cluster editing algorithm to enforce transitivity without adding many spurious links, leading to overall improved performance.

\end{abstract}

\section{Introduction}

Semantic web technology is increasingly used by domain experts, such as historians, to answer important questions in their respective fields. In order to answer many of these questions, access to information that is present in large knowledge graphs is required. However, due to the independent nature of the institutions that govern the datasets, their respective knowledge graphs often use different URI's to refer to the same real-world objects. In order to access the information that is present for an object in large knowledge graphs, automated methods for identifying and linking duplicate entities in the knowledge graphs are required. Linking duplicate entities in the semantic web literature is also known as entity matching~\cite{kopcke2010frameworks}\cite{papadakis2016comparative}\cite{simonini2019scaling}.

The problem of entity matching is to find all possible links between entities that represent the same real-world entity. The result of entity matching can be seen as a set of entity pairs, each pair interpreted as the same-as relation. However, the identified set of pairs may fail to satisfy some structural properties, in particular transitivity, that are expected from the same-as relation. We assume that the entity pairs generated by an entity matching algorithm are used by, for instance, a reasoner in a SPARQL engine. This reasoner will consider the transitive closure of the supplied entity pairs, thereby possibly concluding many spurious entity pairs. These possibly distinct entity pairs are a problem for when, for instance, domain experts subsequently use these pairs in their SPARQL queries. The answers to these queries are then likely to contain errors, making a proper interpretation of the results difficult. A considerable difficulty with entity matching is that the total number of possible entity pairs is much larger than the number of actual (duplicate) entity pairs, also known as the problem of skewness~\cite{weiss2004mining}~\cite{al2011survey}. This extreme skewness can cause false positive results to overwhelm the true positives, even for highly accurate classifiers. This has caused many other works to use ranking techniques, and their associated metrics, to sort the possible entity pairs with some similarity measure, where duplicate entity pairs are expected to appear on top of the ranking \cite{sun2017cross}\cite{zhu2017iterative}\cite{chen2018co}\cite{trisedya2019entity}.

Our contribution is the application of cluster editing in order to create a set of entity pairs that is closed under transitivity, without introducing spurious pairs. This is achieved by first addressing the problem of skewness with the use of an efficient $k$ nearest neighbor algorithm that determines for each entity the $k$ potential duplicates using a given similarity measure. This yields a set of links between entities which are (relatively) likely to be duplicates. These pairs, considered as links between entities, are then used to construct a graph consisting of a number of connected components. These connected components are in turn used as input for the cluster editing algorithm, yielding a set of disjoint cliques (clusters) each of which contains duplicates that refer to one and the same real-world object. The links (i.e., entity pairs) that make up these clusters together constitute a linkset, which can afterwards be used in downstream tasks such as reasoning over the same-as relation in different knowledge graphs.

We set up a number of experiments and show that the application of cluster editing, compared to the transitive closure of likely links (e.g. entity pairs that are top ranked), always results in a 
linkset that contains fewer distinct entity pairs (i.e. they have a higher precision) while at the same time retaining duplicate entity pairs (i.e. recall is not lowered). The experiments are performed on some semi-synthetic datasets that are generated by introducing duplicates in an existing dataset in a controlled manner. This results in a range of different cluster distributions, where we measure the effects of the number of clusters and different cluster sizes. 

The structure of this paper is as follows. First, we present our method for finding duplicates in knowledge graphs by means of a cluster editing technique. We then explain the experimental setup, in particular, how the semi-synthetic data is generated and the experiments are performed. The results of the experiments are then discussed and evaluated in detail. Finally, we present some related work and conclude the paper with some future research directions.

\section{Method}

A knowledge graph $G = (E, L, R)$ consists of a set of entities $E$, of literals $L$, and of relations $R$. 
The objective of entity matching in our setting is, given a set of knowledge graphs $\{G_1, ..., G_n\}$ with $G_i=(E_i,L_i,R_i)$, 
to find the subset of pairs of duplicate entities $\mathcal{L}$, where

\begin{eqnarray*}
    E & = & \bigcup_{i=1}^n E_i, \\
    P & = & E \times E \text{, and} \\
    \mathcal{L} & \subseteq &  P. 
\end{eqnarray*}

We assume we have a distance measure for pairs of entities in $E$ and a classifier that outputs the fitted probability $p_{ij}$ that entities $e_i$ and $e_j$ are duplicates. Since the set $\mathcal{L}$ is to represent the set of duplicate pairs (i.e., $\mathcal{L}$ is to be interpreted as representing the same-as relation), we further require that $\mathcal{L}$ must be closed under transitivity. In the rest of this paper, we use terms `entity pairs' and `links' interchangeably. We also use the term `linkset' to refer to the set of entity pairs that link duplicate entities.     


\begin{figure}
    \centering
    \includegraphics[width=\textwidth]{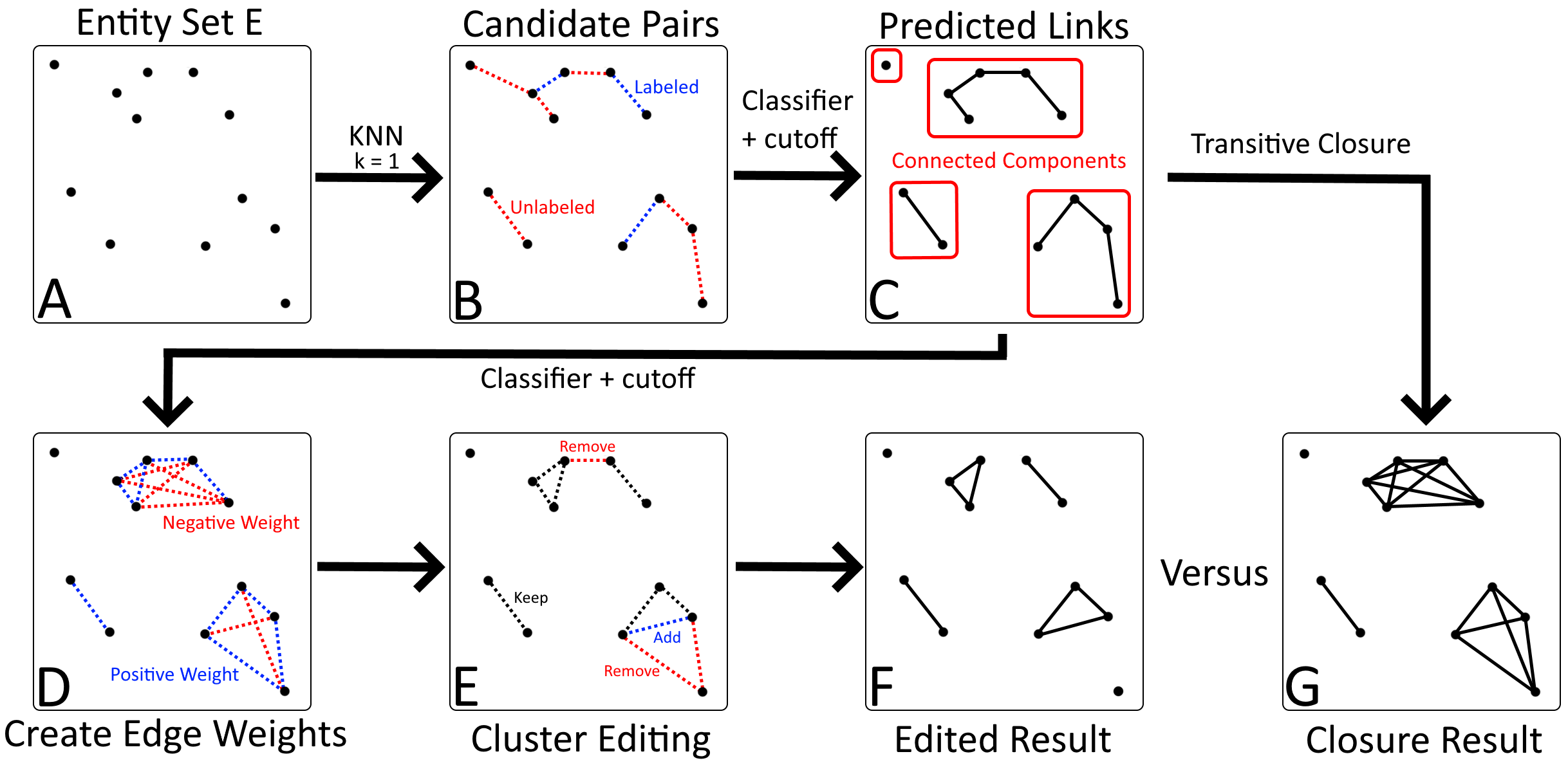}
    \caption{An overview of the linkset generation process.}
    \label{fig:flow}
\end{figure}

Before we discuss the different steps of the process of finding duplicates in detail, we first give a broad overview of it (see figure~\ref{fig:flow}). 
We start with a set of entities $E$, some of which may be duplicates,
and use Euclidean distance to measure their proximity (see panel A of figure~\ref{fig:flow}). 
Let $N_k(i)$ denote the index set of the $k$ nearest neighbors of $e_i$.
For each entity $e_i$, we make $k$ candidate pairs $(e_i,e_j)$, $j \in N_k(i)$.
The dotted lines in panel B indicate the candidate pairs for $k=1$.
Moreover, we assume that a (small) subset of these candidate pairs is labeled by a domain expert (the blue lines in panel B). The labeled pairs are used
to train a probabilistic classifier. This classifier is used to determine, for each candidate pair $(e_i,e_j)$, the fitted probability
$p_{ij}$ that $e_i$ and $e_j$ are duplicates. Depending on the features used by the classifier, and its complexity, the fitted probabilities need not be proportional
to the distance between entities (and in the example in figure~\ref{fig:flow} they are not). We do however assume that the features used by the classifier are
symmetric so that $p_{ij}=p_{ji}$, and therefore we can indeed regard a pair of entities as unordered.
We then use a cut-off value $\theta$ so that if $p_{ij} > \theta$, then $e_i$ and $e_j$ are predicted to be duplicates (or linked; see panel C). 
This ``raw outcome" of the pairwise classifier may however violate the transitivity constraint. 
An ad-hoc way to solve this issue, is to take the
transitive closure of the links predicted by the classifier (panel G). This way of restoring transitivity obviously never removes any links, but
can only add new links. A more principled method to restore transitivity is to use the cluster editing technique. 
Here, we compute a weight $w(i,j)$ for each pair of entities 
$(e_i,e_j)$ within the same connected component (regardless of whether it is a candidate pair or not), 
such that $w(i,j)$ is positive if $p_{ij} > \theta$, and negative otherwise (panel D).
If $w(i,j)$ is positive (negative), a link between $i$ and $j$ is provisionally assumed to be present (absent). 
The resulting set of links may however again violate the
transitivity constraint. Cluster editing is used to restore transitivity by adding and/or removing links in such a way that 
the total score
\[
\sum_{(i,j)}  w(i,j)x_{ij}
\]
is maximized, where $x_{ij}=1$ if a link between $i$ and $j$ is present in the solution, and $x_{ij}=0$ otherwise (panels E and F). 

\subsection{Decreasing Rarity of Duplicate Pairs}
We want to determine the subset $\mathcal{L} \subseteq P$ that contains the duplicate pairs. 
However, the vast majority of pairs in $P$ do not link duplicate entities, so duplicate pairs are very rare~\cite{al2011survey}. 
We reduce the degree of rarity by creating two subsets $P_1 \subseteq P$ and $P_2 = P - P_1$, where $P_1$ is expected to be much smaller in size and to have a much higher occurrence of duplicate pairs.

We construct $P_1$ by, for every entity in $E$, taking its $k$ nearest neighbors from $E$, 
where the value of $k$ is tailored to the specific problem. For example, if we know there can at most be 1 duplicate for each entity, we set $k$ equal to 1. The problem of $k$ nearest neighbor search has been well researched and calculating the $k$ nearest neighbors for all entities can be achieved in $O(k |E| + |E|^{\frac{3}{2}})$ time. Finally, we name $P_1$ the \textit{candidate pairs}, as these are pairs which have a (relatively) high likelihood of being duplicates. 
Duplicate pairs are even more rare in $P_2$ than in $P$, so we tentatively assume that all pairs in $P_2$ refer to distinct entities.

\subsection{Creating a Tentative Linkset}
A (small) subset is randomly selected from the set of candidate pairs, and labeled by a domain expert as `duplicates' or `distinct'. 
This labeled set is used to train a probabilistic classifier.
The trained classifier is used to compute, for each unlabeled candidate pair $(e_i,e_j)$,
the probability $p_{ij}$ that they are duplicates. 
If $p_{ij} > \theta$, then $(e_i,e_j)$
is added to the tentative linkset $\mathcal{L}'$. In addition, the pairs that were labeled as duplicates by the domain expert are added to $\mathcal{L}'$.
The resulting linkset may already be of high quality but still has an issue: the classifier has not considered transitivity when labeling pairs. Submitting $\mathcal{L}'$ to be used by an of the shelf RDF reasoner can result in undesirable effects, such as the transitive closure being computed and used, possibly linking many unrelated entities by long chains.

\subsection{Repairing the Tentative Linkset}
In this section we describe how $\mathcal{L}'$ is modified to remove any transitivity violations. These violations can be fixed by either adding or removing links. The problem then becomes modifying $\mathcal{L}'$ with the minimum number of link deletions and insertions, such that the resulting linkset contains no transitivity violations. This is exactly a problem for which cluster editing~\cite{bocker2013cluster}\cite{bocker2011exact} can be used. 
To conform to the problem representation that is common for cluster editing, we construct a graph that
contains an edge between entities $e_i$ and $e_j$ whenever they occur as a pair in $\mathcal{L}'$. 
Next, we determine the connected components of this graph. 
Each connected component is solved separately using cluster editing. In doing so we substantially reduce the computational complexity, 
at the slight risk of missing potential links between non-candidate pairs that ended up in different connected components. 
Using the classifier, the probability of a link is calculated for every possible pair inside the same connected component, 
regardless of whether it is a candidate pair or not. Like before, a link is provisionally assumed to be present if $p_{ij} > \theta$, and
absent otherwise. Probability $p_{ij}$ is then converted to a weight:
pairs for which $p_{ij} < \theta$ get a negative weight and thus an insertion cost, while pairs for which $p_{ij} > \theta$ get a positive weight, 
and thus a deletion cost. How exactly the weights are computed using $p_{ij}$ and $\theta$ is explained in section \ref{subsection:pair-weights}.

An exact solution to weighted cluster editing can be found with Integer Linear Programming (ILP).  
The ILP formulation for cluster editing~\cite{grotschel1989cutting} is  

\begin{equation} \label{eqn:ILP-min}
    \min_{\bm{X}} \ \sum_{(i,j) \in \Lambda} w(i,j) \ - \sum_{1 \le i < j \le n} w(i,j)x_{ij} 
\end{equation}
\begin{eqnarray} 
    \text{subject to} & \ \ + x_{ij} + x_{jk} -  x_{ik} \leq 1  & \text{for all} \ 1 \le i < j < k \le n,\label{eq:ILP-constraint1} \\
    & \ \ + x_{ij} - x_{jk} +  x_{ik} \leq 1  & \text{for all} \ 1 \le i < j < k \le n,\label{eq:ILP-constraint2} \\
    & \ \ - x_{ij} + x_{jk} +  x_{ik} \leq 1  & \text{for all} \ 1 \le i < j < k \le n,\label{eq:ILP-constraint3} \\
    & x_{ij} \in \{0, 1\} & \text{for all} \ 1 \le i < j \le n \label{eq:ILP-constraint4}
\end{eqnarray}

Here $w(i,j)$ is the weight given to the pair $(e_i,e_j)$, 
$\Lambda$ denotes the set of  pairs with a positive weight,
and $x_{ij} = 1$ when a link between $e_i$ and $e_j$ 
is part of the solution, and zero otherwise. $\bm{X}$ denotes the strictly upper triangular $n \times n$ matrix containing all $x_{ij}$ values,
where $n$ is the total number of entities in the connected component.
Constraints~(\ref{eq:ILP-constraint1})~-~(\ref{eq:ILP-constraint3}) together ensure that the solution satisfies transitivity.
The first sum in the objective function~(\ref{eqn:ILP-min}) represents the quality of the unconstrained optimal solution in which only links with positive
weights are included. 
Since this is term does not depend on the decision variables $x_{ij}$, the objective function can be simplified to: 

\begin{equation} \label{eqn:ILP-max}
    \max_{\bm{X}} \ \sum_{1 \le i < j \le n} w(i,j)x_{ij} 
\end{equation}

Applying cluster editing to each connected component is expected to have the following effects:
\begin{enumerate}
    \item Components that are only weakly connected tend to be discarded, i.e. the optimal solution is to have no links at all. 
    This deals with the problem of long chains of links.
    \item Clusters of entities which are weakly interconnected tend to be split up.
\end{enumerate}
Additionally, components that are too large can not be computed in reasonable time and are discarded. These components have a high probability of consisting of multiple cluster of entities that are very hard to distinguish. Removing these components will lower recall but increase precision of the resulting linkset, 
since any valid links between entities in those components are not added, but at the same time we refrain from adding many invalid links.


\section{Experimental Setup}

For our experiments we make use of semi-synthetic data in order to have precise control over important properties of duplicate entities. This means that we start with a real-world knowledge graph and create duplicate entities in the knowledge graph ourselves, such that the true linkset $\mathcal{L}$ is known in advance. Using this method allows us to compare the linkset $\mathcal{L}'$ computed by our approach to the true linkset $\mathcal{L}$ in order to determine the exact precision and recall values. Furthermore, generating the duplicates ourselves allows for more fine tuning of cluster size distributions, which can then be compared in how they affect performance in precision and recall.

\subsection{Data Generation}
The source of our data is an RDF version of Ecartico\footnote{\url{http://www.vondel.humanities.uva.nl/ecartico}}, a comprehensive dataset with biographical data about, among others, painters, engravers and book sellers. These people worked in the Low Countries (the Southern as well as Northern Netherlands; later `the Dutch Republic') at the time of the sixteenth and seventeenth century. Ecartico satisfies a number of properties important for our purposes: First, it is actively curated so we can be sure there are no duplicate records of persons present, which would be a source of potential errors we cannot account for. Second, person records have a sufficient number of properties to be split across multiple subsets while still maintaining some overlap, i.e. allowing for properties that occur in more than one subset. This is a property that our embedding creation method exploits.

From the Ecartico graph, we have constructed a new graph containing all \texttt{schema:Person} entities that have values for the properties \begin{itemize*}
    \item \texttt{schema:name},
    \item \texttt{schema:workLocation}, and 
    \item \texttt{schema:hasOccupation}.
\end{itemize*}
For each of these entities, we also copy the property-value pairs for 
\begin{itemize*}
    \item \texttt{schema:birthDate}, \\
    \item \texttt{schema:deathDate}, 
    \item \texttt{schema:birthPlace}, and 
    \item \texttt{schema:deathPlace} 
\end{itemize*}
when they are present. This resulted in a graph with, including \texttt{rdf:type} for the class \texttt{schema:Person}, eight properties, all centered around that one RDF class. 

\begin{figure}
    \centering
    \includegraphics[width=\textwidth]{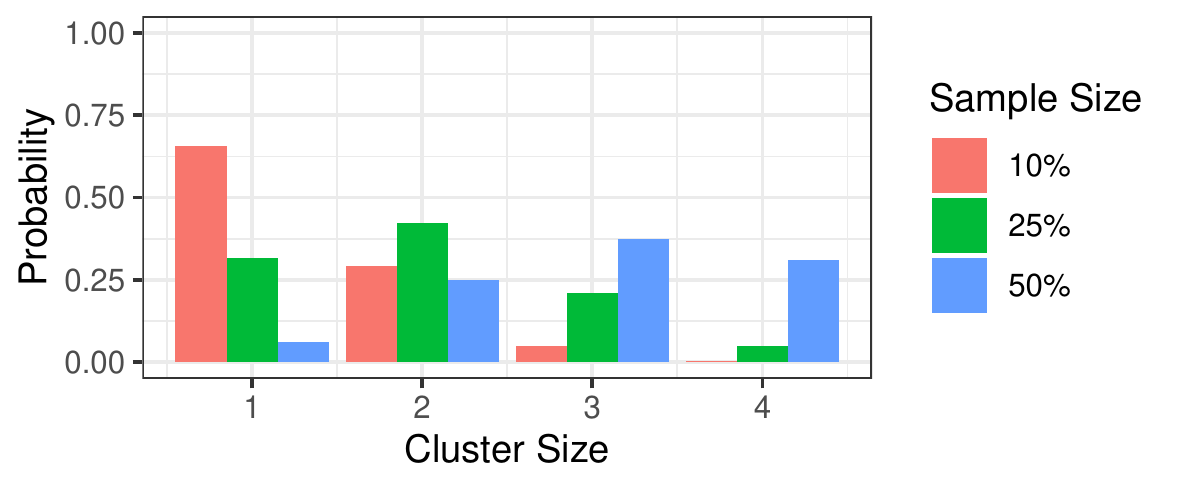}
    \caption{Generated probability distributions of entity clusters of size 1 to 4 in the synthetic data. The values of a color sum to one.}
    \label{fig:cluster-size-distribution}
\end{figure}

This new graph is then in turn split up into four subgraphs, where every entity is present, with an identical URI, in all four subgraphs. However, each subgraph contains different properties, with some overlap. For example, multiple subgraphs may contain the property \texttt{schema:name}, while only a single subgraph contains the property \texttt{schema:birthDate}. Then we introduce duplicates by, independently for each subgraph, uniformly sampling a percentage of entities and altering their respective URI's by appending the subgraph number. For example, when sampling entities from the third subgraph, an entity with URI \texttt{www.data.uu.nl/1234} will be modified to \texttt{www.data.uu.nl/1234/3}. We create three cluster distributions $\mathcal{D}_{10}$, $\mathcal{D}_{25}$ and $\mathcal{D}_{50}$, which are deteremined by the percentage of entities that is sampled, which can be either $10$, $25$ or $50$ percent. Figure \ref{fig:cluster-size-distribution} shows the resulting distributions of clusters, where, for instance with $10$\%, we expect most entities to be in a cluster of size 1, meaning they were not duplicated. When modifying $50$\%, of entities, however, this results in most entities being duplicated at least once, and the largest proportion being duplicated twice. In this case non-duplicate entities are in the minority, making the entity matching problem considerably harder. 

Table \ref{tab:dist-info} shows, for each cluster distribution, the total number of non unique entities $|E|$ and the skew present in both the set of all pairs $P = E \times E$ as well as the candidate pairs. It is clear that the candidate pairs have a much higher occurrence of duplicate pairs compared to all possible pairs. Note that especially in the case of $\mathcal{D}_{10}$, most entities are not duplicated and therefore we expect a lower ratio.

\setlength{\tabcolsep}{0.5em} 
{\renewcommand{\arraystretch}{1.2}
\begin{table}[H]
    \centering
    \caption{Total number of entities $|E|$ and the ratio of duplicate pairs to all possible pairs $P$, and the candidate pairs, that are used in experiments.}
    \begin{tabular}{lccc}
    \hline
    Cluster Distribution & $|E|$ & $P$ class ratio & Candidate pairs class ratio \\
    \hline
    $\mathcal{D}_{10}$  & 4553 & $\frac{1155}{10362628} = 0.011\%$  & $\frac{1049}{10430} = 10.06\%$  \\
    $\mathcal{D}_{25}$  & 6128 & $\frac{3283}{18773128} = 0.017\%$  & $\frac{2621}{13320} = 19.68\%$  \\
    $\mathcal{D}_{50}$  & 8756 & $\frac{7907}{38329390} = 0.021\%$  & $\frac{5238}{17275} = 30.32\%$  \\
    \hline
    \end{tabular}
    \label{tab:dist-info}
\end{table}
}

Once the clusters have been generated, we make use of our embedding creation method~\cite{baastailored} to generate an embedding for each cluster distribution, yielding three embeddings. These embeddings each have $100$ dimensions and contain all relevant entities for a particular cluster distribution. Note that any method that is able to embed entities from multiple knowledge graphs into a single embedding can be used.

\subsection{Entity Pairs}\label{subsection:pair-weights}
The nearest neighbors that form the candidate pairs are calculated with the Approximate Nearest Neighbors\footnote{\url{http://www.cs.umd.edu/~mount/ANN}} (ANN) library, using Euclidean distance. The main efficiency bottleneck for nearest neighbor search is the number of dimensions in the embedding, where, as the number of dimensions grows, the search increasingly resembles a brute force linear search per entity. The ANN library allows for the use of an error bound in order to increase efficiency. For our experiments we have set the error bound to zero, resulting in exact nearest neighbor search. It is possible, however, to sacrifice some performance in finding the relevant candidate pairs in exchange for a faster search.

To classify the entity pairs we have trained an elastic net logistic regression (LR) classifier and a support vector machine (SVM) with a radial kernel. As features for entity pairs we use their cosine similarity and the Hadamard product. The Hadamard product is the element-wise product of two vectors, and thus generates as many features as the number of dimensions in the embedding. All models are tuned using a random hyperparameter search. 

For each cluster distribution, we randomly sample $100$ candidate pairs in case of cosine similarity, and $300$ pairs in case of Hadamard product (see table \ref{tab:results} for an overview of the combinations). This is an acceptable number of pairs for a domain expert to label in one or two days. Then, using the classifiers, we associate with every unlabeled pair $(e_i,e_j)$, a probability $p_{ij}$ that they are duplicates. 
To simulate the inclusion of the domain expert's knowledge, we assign any pair that was in the train set a probability of $1 - \epsilon$ for 
duplicate pairs and a probability $\epsilon$ for distinct pairs, for a near zero value $\epsilon = 10^{-6}$.
This probability is the basis of the weight that will be associated with a potential edge between entities $e_i$ and $e_j$ by the cluster editing algorithm.
The calculated weight $w(i,j)$ associated with the pair of entities $e_i$ and $e_j$ is:
\begin{equation}\label{eqn:weight}
    w(i,j) = \log\left({\frac{p_{ij}}{1-p_{ij}}}\right) - \log\left({\frac{\theta}{1-\theta}}\right)
\end{equation}
The logit transformation of $p_{ij}$ in equation~(\ref{eqn:weight}) introduces a non-linearity in the weights, 
where low probability pairs tend to be excluded and high probability pairs tend to be included in the clusters
by the cluster editing algorithm.
Subtracting the logit transformation of $\theta$ makes sure that $w(i,j)$ is positive if $p_{ij} > \theta$, and
$w(i,j)$ is negative otherwise.

Note that setting $\theta$ to some value near zero will cause even low confidence candidate pairs to be initially considered as linked, 
thus increasing the number and size of  connected components. For values of $\theta$ close to one, the opposite is the case: only few pairs with high probability are 
initially considered as linked and, due to the logit transformation, there is a very high cost associated with linking low confidence pairs. 
 

\subsection{Baseline per Cluster Distribution}
For each cluster distribution we compare the performance of two linksets:
\begin{enumerate}
    \item \textit{Closure linkset}: As a baseline, consider as links only the candidate pairs where $p_{ij} \geq \theta$, and then report the transitive closure of these links as a linkset.
    \item \textit{Edited linkset}: Use as a linkset all clusters returned by cluster editing.
\end{enumerate}

Due to the NP-hardness of ILP we are only able to perform cluster editing for connected components containing 50 entities or fewer. This means that, for both the Closure and edited linksets, we discard all components larger than 50 entities. This action can cause recall to be low for very small values for $\theta$, as a small $\theta$ tends to generate many large connected components, which in turn get discarded and left out of the linkset (see next section for an elaborated analysis of the experimental results). We have chosen to unilaterally discard these components to create a fair comparison where all things are equal, except the use of cluster editing and transitive closure.

For all experiments we use the number of nearest neighbors $k = 3$ to generate the candidate pairs. This value for $k$ generates least 3 candidate pairs per entity, which is enough to capture the largest possible clusters (of size $4$) in our experiments.

\section{Results}

Table \ref{tab:results} shows the results of our experiments. We denote the application of transitive closure with the subscript $TC$ and the application of cluster editing with the subscript $CE$. For every value of $\theta \in (0, 1)$ (in steps of 0.01) we generate a precision, recall and associated F-score. It is our experience that a low precision has a larger negative impact (than low recall) on the performance of downstream systems such as SPARQL engines. We therefore use the F$\frac{1}{2}$-score, which weights precision twice as heavy as recall, to measure the overall performance. There is, for a certain cutoff value, an interaction between the choice of candidate pairs, the subsequent connected components, and the result of transitive closure and cluster editing. Therefore precision (and recall) do not necessarily increase (and decrease) monotonically, and metrics such as the area under the precision-recall curve are not applicable without modification. Therefore we average the F$\frac{1}{2}$-score for all values of $\theta$ (100 values between 0 and 1) to denote the performance of a given combination of cluster distribution, classifier and features. 

In all cases we observe that the application of cluster editing improves the resulting linkset over the application of transitive closure. Highlighted in bold are the best mean F$\frac{1}{2}$-scores, attained by the support vector machine classifier using cosine similarity as the feature, and the best overall F$\frac{1}{2}$-scores. Furthermore, the optimal value for $\theta$ is in most cases reduced when cluster editing is applied, suggesting that a more lenient cutoff can be used, while at the same time improving performance.

\begin{table}
    \centering
    \caption{The result of our experiments. A comparison of the mean and maximum F$\frac{1}{2}$-scores, and associated value for $\theta$ of the maximum, per classifier and cluster distributions. Best mean and maximum F$\frac{1}{2}$-scores are highlighted in bold.}
    \begin{tabular}{lccccccc}
      & Train Size & Mean$_{TC}$ & Mean$_{CE}$ & Max$_{TC}$  & Max$_{CE}$ & $\theta_{TC}$ & $\theta_{CE}$  \\
     \hline
     \multicolumn{8}{c}{Logistic Regression - Cosine Similarity} \\
     \hline
     $\mathcal{D}_{10}$ & 100 & 0.46 & 0.50 & 0.69 & \textbf{0.70} & 0.43 & 0.51 \\
     $\mathcal{D}_{25}$ & 100 & 0.37 & 0.41 & 0.62 & \textbf{0.64} & 0.40 & 0.35 \\
     $\mathcal{D}_{50}$ & 100 & 0.38 & 0.44 & 0.62 & \textbf{0.64} & 0.51 & 0.41 \\
     \hline
     \multicolumn{8}{c}{Logistic Regression - Hadamard Product} \\
     \hline
     $\mathcal{D}_{10}$ & 300 & 0.29 & 0.42 & 0.41 & 0.51 & 0.92 & 0.89\\
     $\mathcal{D}_{25}$ & 300 & 0.38 & 0.50 & 0.50 & 0.59 & 0.76 & 0.56 \\
     $\mathcal{D}_{50}$ & 300 & 0.32 & 0.42 & 0.43 & 0.51 & 0.73 & 0.54 \\
     \hline
     \multicolumn{8}{c}{Support Vector Machine - Cosine Similarity} \\
     \hline
     $\mathcal{D}_{10}$ & 100 & 0.61 & \textbf{0.64} & 0.68 & \textbf{0.70} & 0.72 & 0.83 \\
     $\mathcal{D}_{25}$ & 100 & 0.43 & \textbf{0.51} & 0.62 & \textbf{0.64} & 0.70 & 0.66 \\
     $\mathcal{D}_{50}$ & 100 & 0.45 & \textbf{0.54} & 0.62 & \textbf{0.64} & 0.78 & 0.67 \\
     \hline
     \multicolumn{8}{c}{Support Vector Machine - Hadamard Product} \\
     \hline
     $\mathcal{D}_{10}$ & 300 & 0.30 & 0.41 & 0.46 & 0.56 & 0.64 & 0.51 \\
     $\mathcal{D}_{25}$ & 300 & 0.31 & 0.44 & 0.47 & 0.57 & 0.72 & 0.58 \\
     $\mathcal{D}_{50}$ & 300 & 0.25 & 0.33 & 0.40 & 0.48 & 0.70 & 0.59
    \end{tabular}
    \label{tab:results}
\end{table}

\begin{figure}[H]
    \centering
    \subfloat[$\mathcal{D}_{10}$, LR - Cosine Similarity]{
        \includegraphics[width=0.5\textwidth]{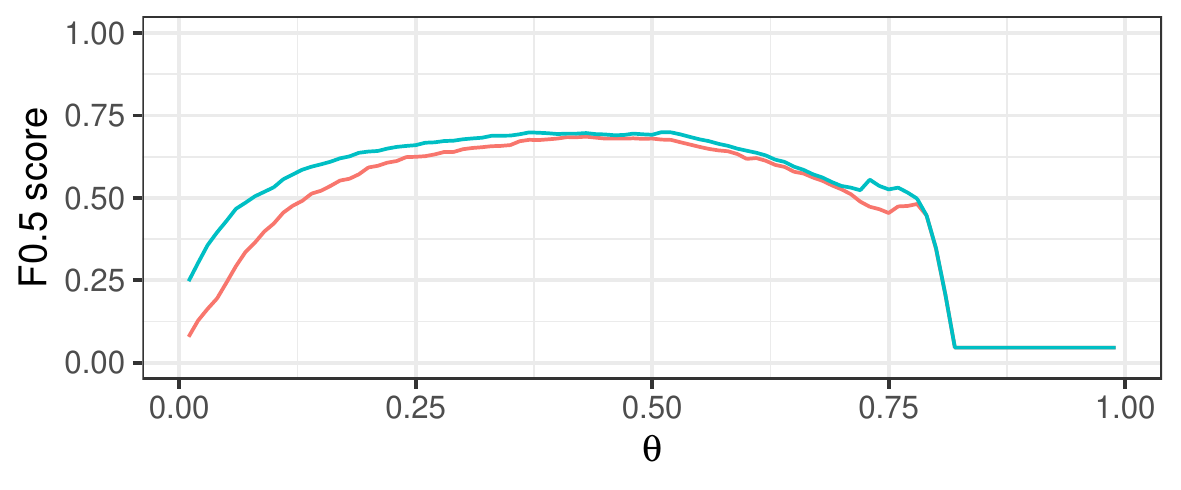}
    }
    \subfloat[$\mathcal{D}_{25}$, LR - Hadamard Product]{
        \includegraphics[width=0.5\textwidth]{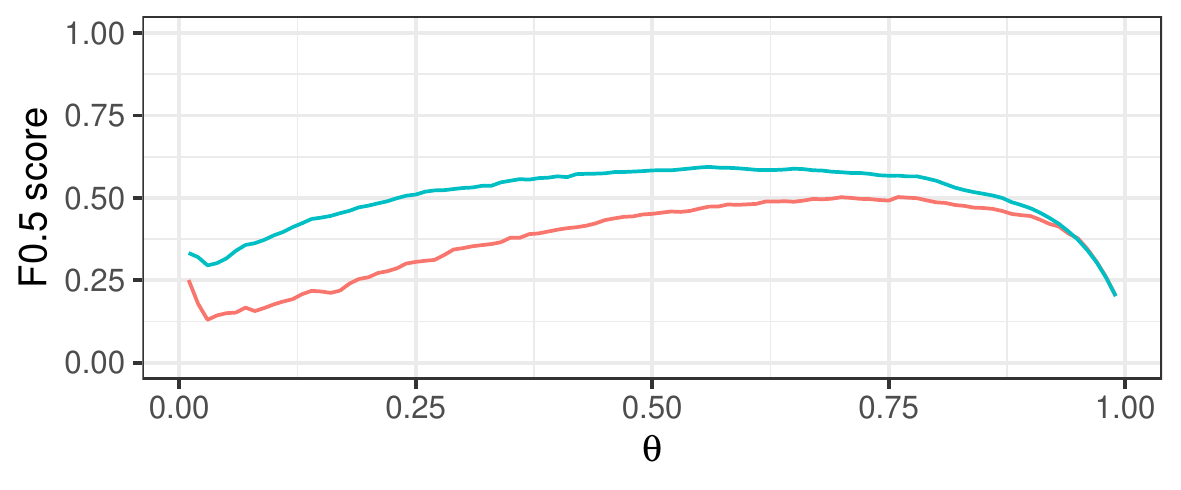}
    }
    
    \subfloat[$\mathcal{D}_{50}$, SVM - Cosine Similarity]{
        \includegraphics[width=0.5\textwidth]{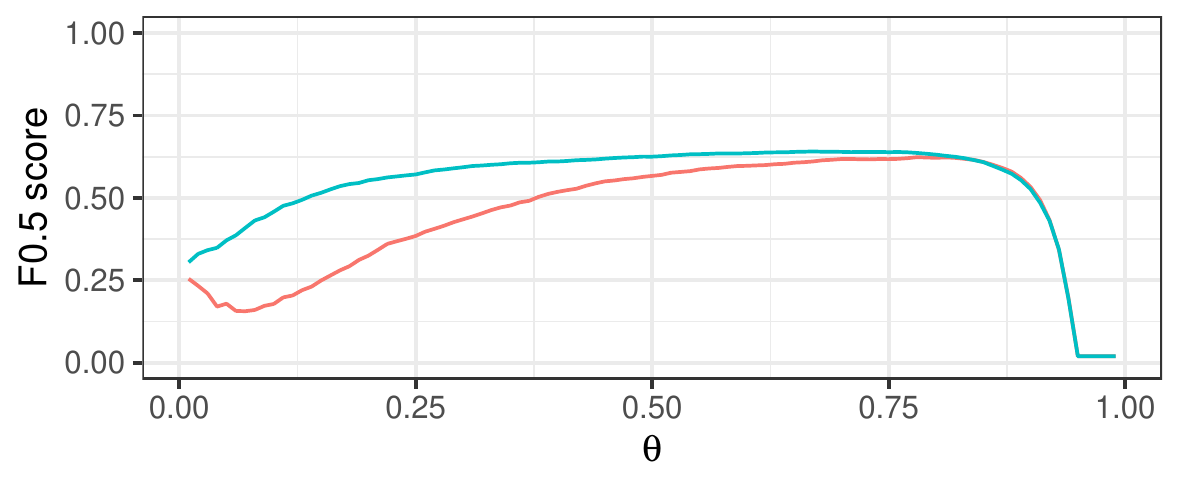}
    }
    \subfloat[$\mathcal{D}_{25}$, SVM - Hadamard Product]{
        \includegraphics[width=0.5\textwidth]{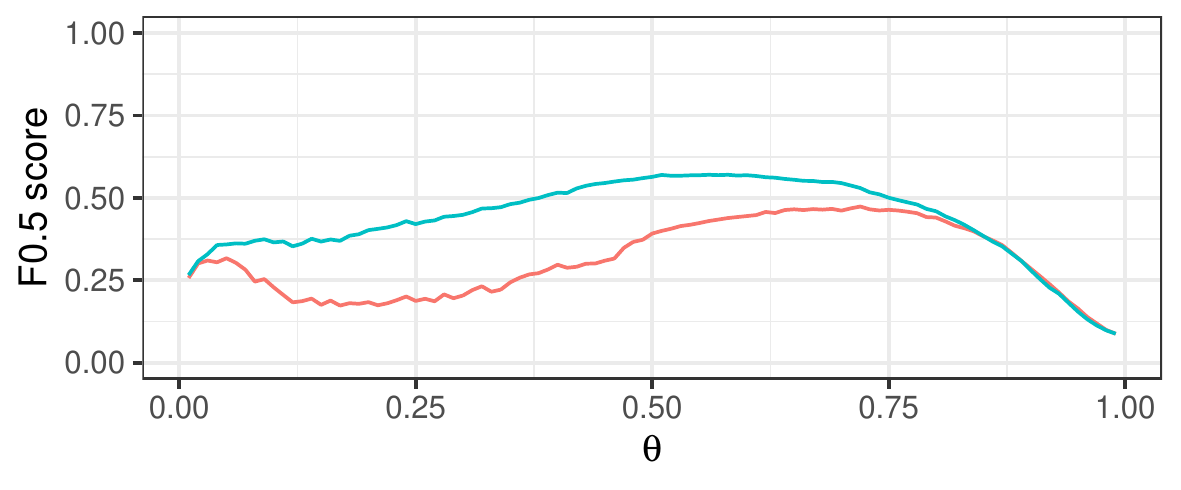}
    }

    \caption{$F_{\frac12}$ scores for Closure linksets (red lines) and Edited linksets (blue lines). Note that the blue lines dominate the red lines.}
    \label{fig:results-fscore}
\end{figure}

We visualize the F$\frac{1}{2}$-scores for a selection of combinations in figure \ref{fig:results-fscore}. The red lines represent the Closure linksets, while the blue lines represent the Edited linksets. The x-axis of each chart is the cutoff value $\theta$, which determines how strict we are in deciding whether a link between entities is valid or not. Observe that the blue lines (Edited linksets) dominate the red lines. This indicates that the cluster editing approach is able to increase F$\frac{1}{2}$-scores most for the low value of $\theta$ ($\theta < 0.5$), where the application of transitive closure tends to introduce many spurious links.

\begin{figure}[H]
    \centering
    \subfloat[$\mathcal{D}_{10}$, LR - Cosine Similarity]{ \label{fig:results-recall-a}
        \includegraphics[width=0.5\textwidth]{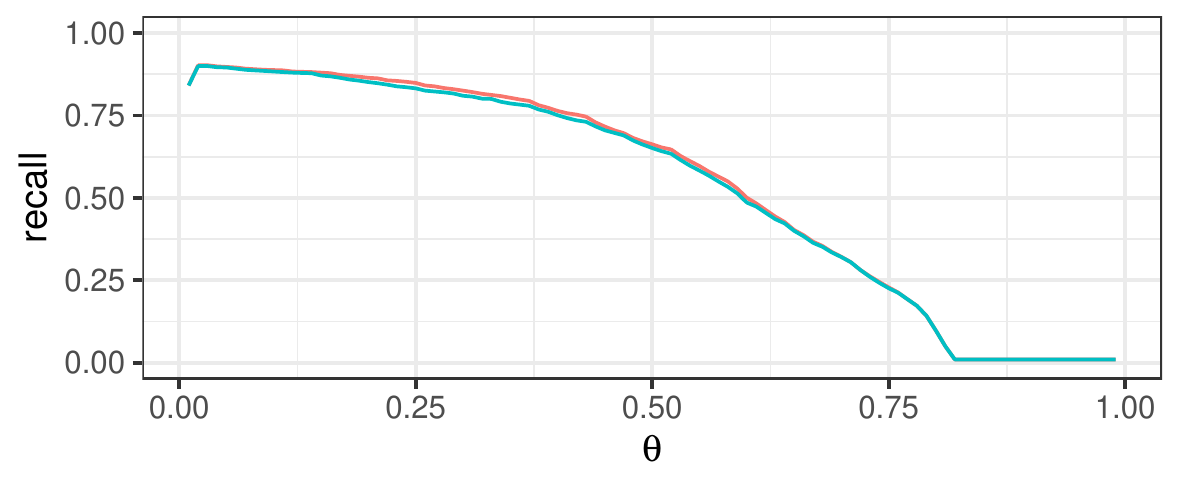}
    }
    \subfloat[$\mathcal{D}_{50}$, SVM - Cosine Similarity]{ \label{fig:results-recall-b}
        \includegraphics[width=0.5\textwidth]{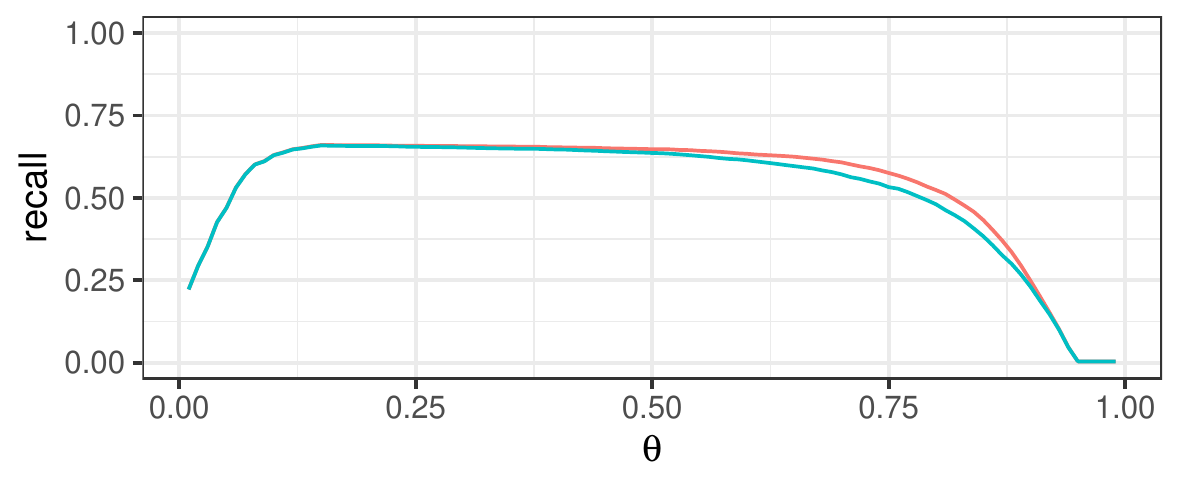}
    }
    \caption{Recall for Closure linksets (red lines) and Edited linksets (blue lines). Note that both lines are nearly identical.}
    \label{fig:results-recall}
\end{figure}

From our experiments we observe that very low values of $\theta$ can have low recall (as seen in figure \ref{fig:results-recall-b}). This is because the low values of $\theta$ cause large connected components to emerge  that will be discarded before cluster editing is applied. As $\theta$ grows, we are more strict and fewer links are taken into consideration (and this few large connected components), until only the candidate links that were in the train set are left. This phenomenon can be seen in the plateaus in figures \ref{fig:results-recall-a} and \ref{fig:results-recall-b}. Note that both the Closure linkset and the Edited linkset have almost identical recall throughout all possible values for $\theta$.

\begin{figure}[H]
    \centering
    \subfloat[$\mathcal{D}_{25}$, LR - Hadamard Product]{ \label{fig:results-precision-a}
        \includegraphics[width=0.5\textwidth]{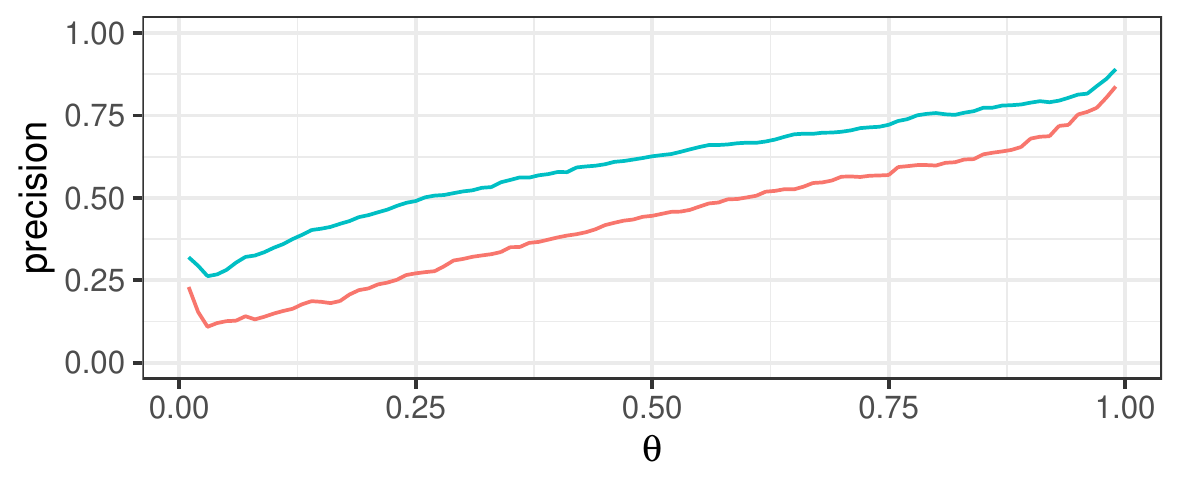}
    }
    \subfloat[$\mathcal{D}_{50}$, LR - Cosine Similarity]{ \label{fig:results-precision-b}
        \includegraphics[width=0.5\textwidth]{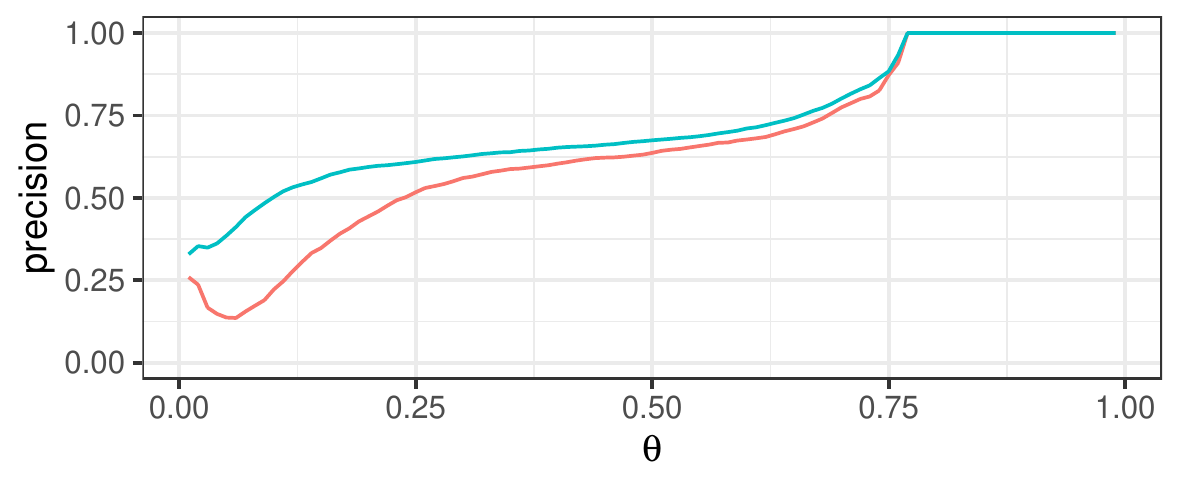}
    }
    \caption{Precision for Closure linksets (red lines) and Edited linksets (blue lines). Note that the blue lines dominate the red lines.}
    \label{fig:results-precision}
\end{figure}

The main contributor to the improvement in F$\frac{1}{2}$-scores is the increase in precision, as can be seen in figures \ref{fig:results-precision-a} and \ref{fig:results-precision-b}. As before with recall, the improvement in precision for low values of $\theta$ is due to the rejection of large connected components, and the plateau for high values of $\theta$ is due to all candidate pairs except those in the train set being rejected. The overall increase in precision is due to cluster editing refraining from adding spurious links which would otherwise be generated by taking the transitive closure.

\begin{figure}[H]
    \centering
    \subfloat[$\mathcal{D}_{25}$, SVM - Cosine Similarity]{ \label{fig:results-size-a}
        \includegraphics[width=0.5\textwidth]{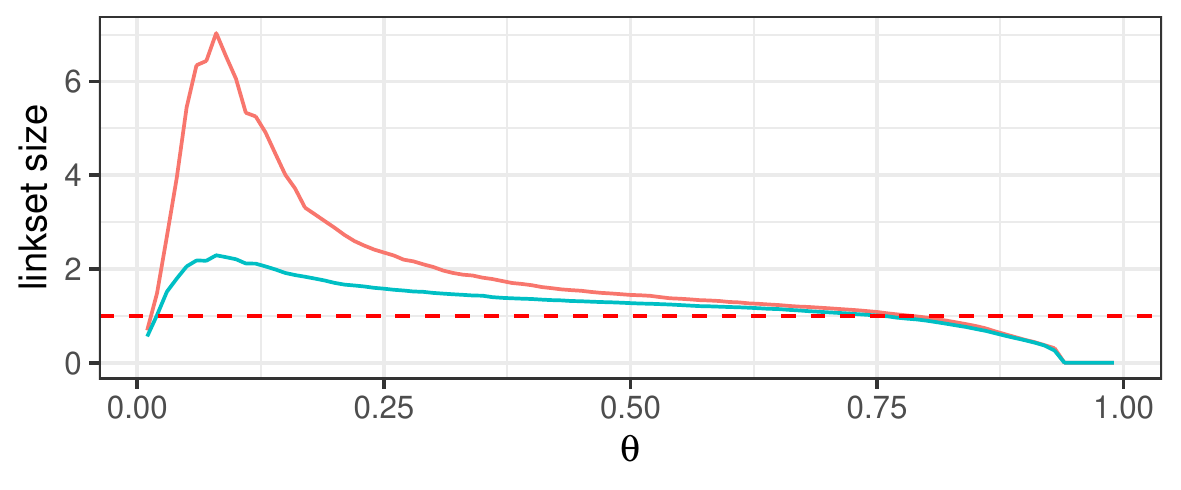}
    }
    \subfloat[$\mathcal{D}_{50}$, LR - Cosine Similarity]{ \label{fig:results-size-b}
        \includegraphics[width=0.5\textwidth]{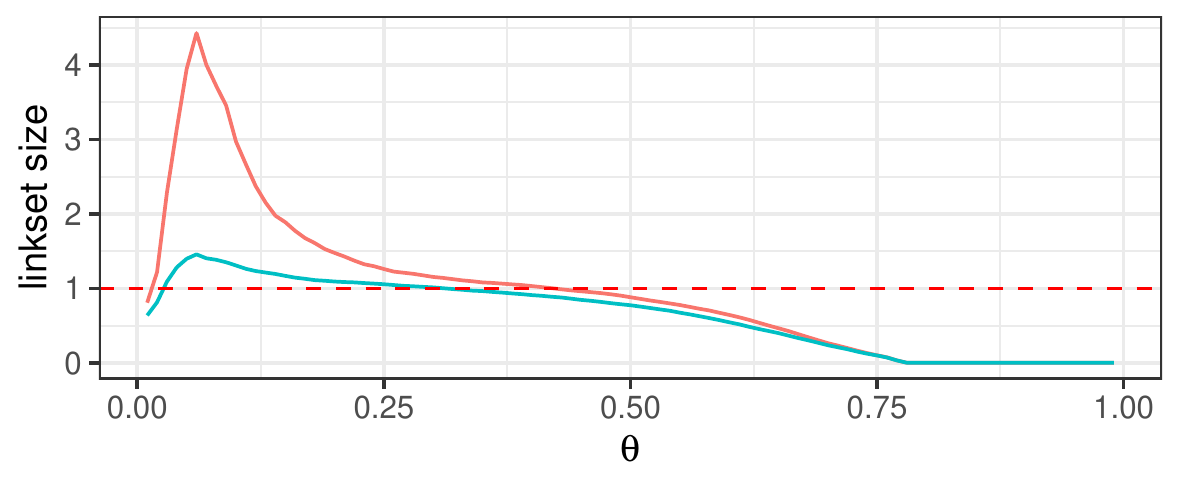}
    }
    \caption{The relative linkset sizes for Closure linksets and Edited linksets.}
    \label{fig:results-size}
\end{figure}

Lastly, in figure \ref{fig:results-size} we show the relative size of linksets, compared to their respective ground truth linksets (red dotted lines). Observe that the Closure linksets (red lines) are consistently much larger than the ground truth linksets, while the Edited linkset sizes tend to be closer to the true linkset sizes over a range of possible $\theta$ values.

\section{Related Work}

In a comprehensive survey of entity matching frameworks, Köpcke et al.~\cite{kopcke2010frameworks} describe entity matching (duplicate identification, record
linkage, entity resolution) as the task of identifying entities (objects, data instances) that refer to the same real-world entity. 
They make no assumptions about the structure of the data (relational tables, RDF graphs, etc.), 
or the way that duplicates are represented (e.g. as pairs or as clusters). 
Several high-level requirements that should be met are listed for entity matching frameworks, notably \textit{effectiveness}: to achieve a high quality match with regard to precision and recall. This is what we have focused on in our work. But also \textit{low manual effort}, which is important in our case as the time of domain experts is expensive and ought to be used as little as possible.

Related to entity matching is entity \textit{alignment}. This term is normally used when two knowledge graphs $G_1$ and $G_2$ are considered and, for every entity in $G_1$, there exists at most one duplicate entity in $G_2$. This specific problem occurs for cases such as matching entities between two KG's of different languages. Examples are the work of Chen et al.~\cite{chen2018co}, who leverage a Wikipedia based dataset to co-train two embedding models, and Trisedya et al.~\cite{trisedya2019entity} who allow for the integration of multiple knowledge graphs into a single space with the use of literals. However, we observe that these works do not use the assumption that the output should be closed under transitivity and report ranking metrics such as hits@k, mean rank (MR) and mean reciprocal rank (MRR). These metrics are appropriate for tasks such as information retrieval, where the presence of relevant results at the top of the ranking is sufficient. However, they do not take into account the potential loss in performance due to the introduction of spurious links when it is  a requirement that the output is closed under transitivity.

Coreference resolution in natural language processing has a similar purpose as entity matching in knowledge graphs. 
It is concerned with deciding which noun phrases in a document refer to the same real world entity.
Finkel and Manning \cite{finkel2008enforcing} use Integer Linear Programming to enforce transitivity by post-processing the predictions of a pairwise classifier for coreference resolution. Their ILP objective is to maximize the log probability of the coreference decisions given the fitted probabilities:
\begin{equation}\label{eq:manning1}
\max_{\bm X} \sum_{i,j} x_{ij} \log p_{ij} + (1-x_{ij}) \log (1-p_{ij})
\end{equation}
where $p_{ij}$ denotes the probability estimate of the pairwise classifier that noun phrases $i$ and $j$ are coreferent, $x_{ij}=1$ if 
$i$ and $j$ are decided to be coreferent, and $x_{ij}=0$ otherwise. The summation is over all possible pairs of noun phrases in the document.
The objective function is maximized with respect to $\bm X$, the matrix containing all coreference decisions.
The constraint to be satisfied is that the coreference decisions must be transitive. 

After expanding and rearranging terms in objective function~(\ref{eq:manning1}),
we obtain objective function:
\begin{equation}\label{eq:manning2}
\max_{\bm X}  \sum_{i,j} \log (1-p_{ij}) + \sum_{i,j} x_{ij} \log \left(\frac{p_{ij}}{1-p_{ij}}\right)
\end{equation}
Since the first sum in objective function~(\ref{eq:manning2}) does not depend on the decision variables $\bm{X}$, it can be ignored. The resulting objective function
is a special case of the cluster editing objective function~(\ref{eqn:ILP-max}), with weights as given in 
equation~(\ref{eqn:weight}), and $\theta = \frac{1}{2}$.
Finkel and Manning \cite{finkel2008enforcing} perform experiments on three benchmark data sets, 
and compare the ILP system to a baseline that takes the transitive closure of the links that are present in the unconstrained optimum. 
As expected, taking the transitive closure leads
to a higher recall, as it only adds coreference links. On the other hand, the ILP system may also remove links, and thereby produce a better precision 
and ultimately, a better F-score. Their experiments show that the ILP system obtained a better F-score than the baseline on all three benchmark data sets.

\section{Conclusion and Future Work}
Domain experts such as historians use information from knowledge graphs, however, these knowledge graphs often use different URI's for identical real-world entities. Therefore, in order to access all the attributes associated with an entity, these URI's have to be linked first by means of a linkset.
%
When the resulting linksets are used by, for instance, a reasoner in a SPARQL engine, the transitive closure is applied. The application of transitive closure may introduce many spurious links, potentially creating large clusters of unrelated entities. The presence of these large clusters can be observed in our results (i.e. in figures \ref{fig:results-size-a} and \ref{fig:results-size-b}).
%
Instead of using the transitive closure, we have applied cluster editing in order to create a set of links that is closed under transitivity, without introducing these spurious links.

We have shown that the application of cluster editing, compared to the transitive closure of likely links, always results in a linkset that contains fewer distinct entity pairs (i.e. they have a higher precision) while at the same time retaining duplicate entity pairs (i.e. recall is not lowered). This is done across a range of different cluster distributions, where we measure the effects of the number of clusters and different cluster sizes.

The NP-Hard ILP formulation of cluster editing limits us to solving only relatively small connected components in reasonable time and with limited computational power. There are, however, heuristic methods which enables larger components to be solved~\cite{rahmann2007exact}. Additionally, it is possible to pre-process the connected components and reduce their size by deducing that the optimal solution would never include, or exclude, some links, given a maximum number of allowable edits. These fixed-parameter algorithms offer an efficient pre-processing that effectively reduces the instance size of the problem and are fast in case a small number of edits is allowed~\cite{guo2009more}\cite{chen20122k}. Using one, or both, of these methods will allow us to consider larger connected components and possibly find additional 'nuggets' of clusters. Due to time limitation, we have postponed experimentation with various cluster editing approaches for future research.

Furthermore, the choice of an optimal value for $k$ is not always obvious, and a different method for choosing candidate pairs can be devised. For example, by examining the distribution of similarities between entities (such as distances), it can be possible to exclude certain pairs from being candidates. In the current method, these pairs would first be considered as candidate pairs, but would then more likely than to be judged as distinct by the classifier or domain expert, and thereby removing them from the connected components.

\bibliographystyle{splncs04}
\bibliography{bibs}

\begin{thebibliography}{10}
\providecommand{\url}[1]{\texttt{#1}}
\providecommand{\urlprefix}{URL }
\providecommand{\doi}[1]{https://doi.org/#1}

\bibitem{al2011survey}
Al~Hasan, M., Zaki, M.J.: A survey of link prediction in social networks. In:
  Social network data analytics, pp. 243--275. Springer (2011)

\bibitem{baastailored}
Baas, J., Dastani, M., Feelders, A.: Tailored graph embeddings for entity
  alignment on historical data. In: Proceedings of the 22nd International
  Conference on Information Integration and Web-based Applications \& Services.
  pp. 125--133. ACM (2020)

\bibitem{bocker2013cluster}
B{\"o}cker, S., Baumbach, J.: Cluster editing. In: Conference on Computability
  in Europe. pp. 33--44. Springer (2013)

\bibitem{bocker2011exact}
B{\"o}cker, S., Briesemeister, S., Klau, G.W.: Exact algorithms for cluster
  editing: Evaluation and experiments. Algorithmica  \textbf{60}(2),  316--334
  (2011)

\bibitem{chen20122k}
Chen, J., Meng, J.: A 2k kernel for the cluster editing problem. Journal of
  Computer and System Sciences  \textbf{78}(1),  211--220 (2012)

\bibitem{chen2018co}
Chen, M., Tian, Y., Chang, K.W., Skiena, S., Zaniolo, C.: Co-training
  embeddings of knowledge graphs and entity descriptions for cross-lingual
  entity alignment. arXiv preprint arXiv:1806.06478  (2018)

\bibitem{finkel2008enforcing}
Finkel, J.R., Manning, C.D.: Enforcing transitivity in coreference resolution.
  In: Proceedings of ACL-08: HLT, Short Papers. pp. 45--48 (2008)

\bibitem{grotschel1989cutting}
Gr{\"o}tschel, M., Wakabayashi, Y.: A cutting plane algorithm for a clustering
  problem. Mathematical Programming  \textbf{45}(1-3),  59--96 (1989)

\bibitem{guo2009more}
Guo, J.: A more effective linear kernelization for cluster editing. Theoretical
  Computer Science  \textbf{410}(8-10),  718--726 (2009)

\bibitem{kopcke2010frameworks}
K{\"o}pcke, H., Rahm, E.: Frameworks for entity matching: A comparison. Data \&
  Knowledge Engineering  \textbf{69}(2),  197--210 (2010)

\bibitem{papadakis2016comparative}
Papadakis, G., Svirsky, J., Gal, A., Palpanas, T.: Comparative analysis of
  approximate blocking techniques for entity resolution. Proceedings of the
  VLDB Endowment  \textbf{9}(9),  684--695 (2016)

\bibitem{rahmann2007exact}
Rahmann, S., Wittkop, T., Baumbach, J., Martin, M., Truss, A., B{\"o}cker, S.:
  Exact and heuristic algorithms for weighted cluster editing. In:
  Computational Systems Bioinformatics: (Volume 6), pp. 391--401. World
  Scientific (2007)

\bibitem{simonini2019scaling}
Simonini, G., Gagliardelli, L., Bergamaschi, S., Jagadish, H.: Scaling entity
  resolution: A loosely schema-aware approach. Information Systems
  \textbf{83},  145--165 (2019)

\bibitem{sun2017cross}
Sun, Z., Hu, W., Li, C.: Cross-lingual entity alignment via joint
  attribute-preserving embedding. In: International Semantic Web Conference.
  pp. 628--644. Springer (2017)

\bibitem{trisedya2019entity}
Trisedya, B.D., Qi, J., Zhang, R.: Entity alignment between knowledge graphs
  using attribute embeddings. In: Proceedings of the AAAI Conference on
  Artificial Intelligence. vol.~33, pp. 297--304 (2019)

\bibitem{weiss2004mining}
Weiss, G.M.: Mining with rarity: a unifying framework. ACM Sigkdd Explorations
  Newsletter  \textbf{6}(1),  7--19 (2004)

\bibitem{zhu2017iterative}
Zhu, H., Xie, R., Liu, Z., Sun, M.: Iterative entity alignment via joint
  knowledge embeddings. In: IJCAI. vol.~17, pp. 4258--4264 (2017)

\end{thebibliography}

\end{document}